# Training of CC4 Neural Network with Spread Unary Coding

Pushpa Sree Potluri

**Abstract**
This paper adapts the corner classification algorithm (CC4) to train the neural networks using spread unary inputs. This is an important problem as spread unary appears to be at the basis of data representation in biological learning. The modified CC4 algorithm is tested using the pattern classification experiment and the results are found to be good. Specifically, we show that the number of misclassified points is not particularly sensitive to the chosen radius of generalization.

**1. Introduction**
The idea behind the corner classification approach [1]-[4] to the training of feedforward neural networks is to classify the outputs of the training samples to the corners of a multi-dimensional cube based on the corresponding inputs. This is seen most clearly if one were to viusalize a three-dimensional binary cube where the corners are to be mapped to different classes (Figure 1). The corner classification approach provides instantaneous training which is very important in many applications. The basic algorithms are termed CC1 through CC4 [5],[6], but there are more advanced algorithms as well that provide flexibility in many applications [7]-[10]. General issues related to implementation and learning may be seen in [14]-[17].

In the basic CC4 algorithm the training samples be presented only once during training which is sufficient to determine the interconnection weights. The CC4 network is a three layered feed forward network which consists of Input layer, Hidden layer, and Output layer. The number of neurons in the input layer is one more than the number of elements in the input vector. The additional neuron being the bias neuron, receives a constant input 1. The number of hidden neurons is equal to the number of training samples. Every hidden neuron corresponds to a training sample in the training set and each hidden neuron is connected to all the input neurons.

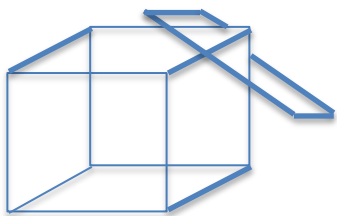
Figure 1. Corner identified by hyperplane

The number of neurons in the output layer is equal to the minimum number of bits required to present the output. The output layer is fully connected; each hidden neuron is connected to each and every output neuron so as to affirm (or negate) membership effectively. The input and output weights to the hidden neurons are assigned using the algorithm related to the hyperplanes.

The CC4 algorithm is now described: The interconnection weight from the bias neuron to the hidden neurons for each training sample is assigned as $r - s + 1$., where $r$ is the radius of generalization and $s$ is the number of 1s present in the corresponding input vector. The value *r* represents the radius of the sphere around the learnt corner in the n-dimensional cube to which the learnt class is generalized. If the point to be generalized is in between two learnt corners, then the effect of learning will persist to half the distance. Points that are exactly half the distance will be generalized by one of the neighboring points in a random fashion.

The weights for the links between input layer and the hidden neurons are assigned based on the following equation:

$$W_i[j] = \begin{cases} 1 & \text{if } x_i[j] = 1 \\ -1 & \text{if } x_i[j] = 0 \\ r-s+1 & \text{if } j = n. \end{cases}$$

The output interconnection weight from each hidden neuron to an output neuron is assigned based on the desired output at the output neuron. If the desired output for the training sample corresponding to the hidden neuron is 1, then the weight assigned is 1 and -1 if the output is 0.

For the CC4 algorithm, unary coding of the data points has been found to be most effective [9],[10]. The reason behind is the fact that the Hamming distance between different points changes gracefully in the unary case as compared to the binary case where points far apart can have a very small Hamming distance as, say, in 000000 and 100000. Therefore, the grid of points on which the data is defined must use many more bits than would be the case either in binary or decimal.

Unary coding has been found to be the basis of learning in certain biological networks [11],[12]. The objective of the present paper is to investigate the use of generalized unary coding [13] in CC4 learning, which was recently presented in the literature. This will open up the applicability of CC4 networks to a much larger world of applications than has been thought. The most important contribution of the paper is to show that misclassification does not change very much as the radius of generalization is increased.

**Unary Coding**
The unary code of a number *n* is represented by *n-1* 1s followed by a zero or by *n-1* zero bits followed by 1 bit.

*n: 00 .... (*n-1 times*) 1*

Table 1: Standard Unary code for numbers 1 to 10

| N | Unary code |
|---|---|
| 0 | 0000000000 |
| 1 | 0000000001 |
| 2 | 0000000011 |
| 3 | 0000000111 |
| 4 | 0000001111 |
| 5 | 0000011111 |
| 6 | 0000111111 |
| 7 | 0001111111 |
| 8 | 0011111111 |
| 9 | 0111111111 |
| 10 | 1111111111 |

In spread unary coding, for a given spread *k*, the number *n* can be represented by *n-1* 0s followed by *k* 1s.

*n: 00… (*n-1 times*) 111..... (*k times*)*

Table 2: Spread Unary code for $k = 3$ for numbers 0 to 10

| n | k = 3 |
|---|---|
| 0 | 000000000000 |
| 1 | 000000000111 |
| 2 | 000000001110 |
| 3 | 000000011100 |
| 4 | 000000111000 |
| 5 | 000001110000 |
| 6 | 000011100000 |
| 7 | 000111000000 |
| 8 | 001110000000 |
| 9 | 011100000000 |
| 10 | 111000000000 |

For any two given numbers, $n_1$ and $n_2$, and for spread *k*, let the distance between them be *d*. The value of *d* depends on the difference between the two numbers. Specifically, as long as the two numbers differ by a value less than the spread, *k*, the distance between them is simply twice the difference between them i.e. $2*(n_2-n_1)$. The distance, hence, is independent of the value of *k*. However, once the numbers differ by a value greater than or equal to *k*, distance between them, now becomes constant at 2*k*. [13]

**Training of patterns**
The original pattern to show training and performance in Reference [8] is shown in Figure 2 (a). The pattern consists of two regions; dividing an 11x16 area into a black spiral shaped region, in which a point is represented as 1 and a point in the white region is represented as 0. Any point in the region is represented by row and column coordinates. These coordinates are encoded using 16–bit unary encoding and fed as inputs to the network. The corresponding outputs are 1 or 0, to denote the region the point belongs to.

The unary code converts numbers 1 to 16 into strings, each of length 16 bits. As an example the number 1 is represented by a string that has fifteen 0s followed by a 1. To represent a point in the pattern, the 16–bit strings of the row and column coordinates are concatenated together. To this, another bit that represents the bias is added and the resulting 33-element vector is given as the input to the network [8].

The training samples are randomly selected points from the two regions of the pattern. Figure 2(b) is an example of a spiral image pattern used for testing the CC4 network. The points marked "#" represents the location is learnt positive, "o" represents the location is learnt negative and all blank spaces represent the locations are not learnt. This matrix is used to train the neural network using the CC4 algorithm.

After the training is done, the network is tested for all 176 points in the 11×16 area of the pattern as the value of *r* is varied from 1 to 4. The results for different values of *r* are shown in the Figures 2(c), (d), (e), and (f). As the value of *r* increases, the network tends to generalize more points belonging to the black region.

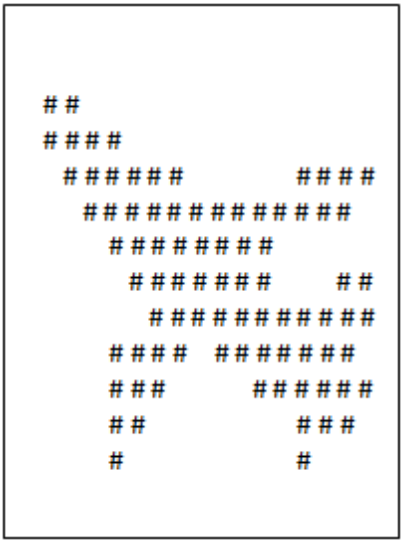

(a). Original Spiral Pattern

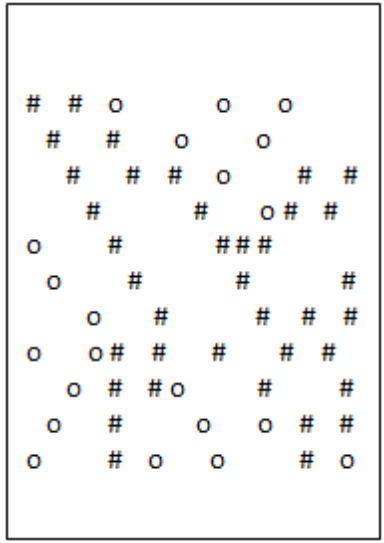

(b). Training Sample

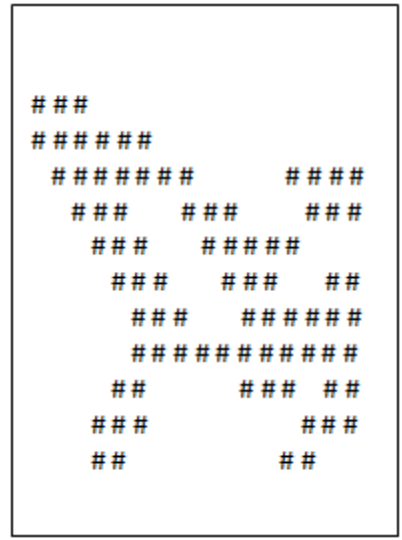

(c). r = 1

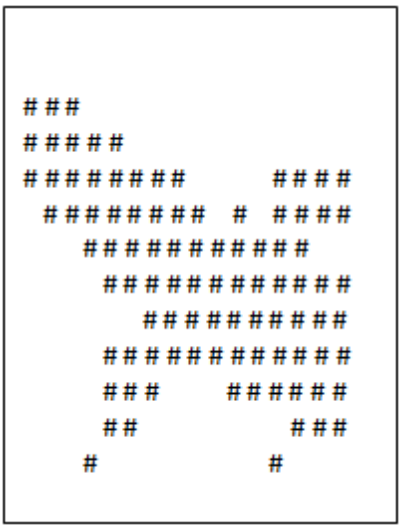

(d). r = 2

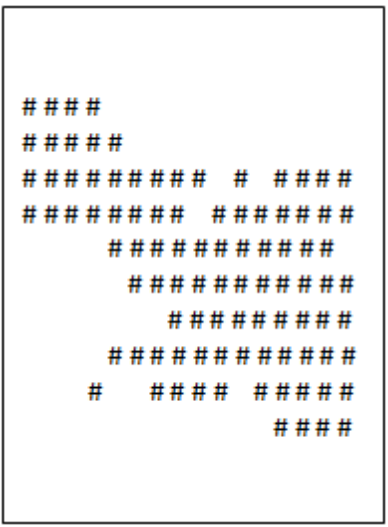

(e). r = 3

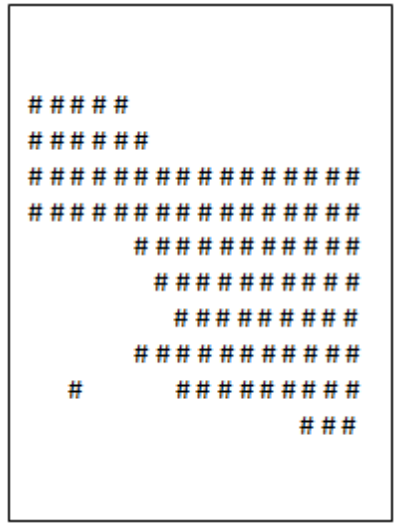

(f). r = 4

Figure 2: Results of Pattern classification using Unary coding

Table 3: No. of points classified/misclassified in the pattern using unary coding

|  | r = 1 | r = 2 | r = 3 | r = 4 |
|---|---|---|---|---|
| Misclassified | 31 | 22 | 36 | 46 |
| Classified | 145 | 154 | 140 | 130 |

We find that in standard unary coding the misclassification rate falls from r=1 to r=2 and then climbs back up. Clearly r=2 is the optimal choice because beyond it we experience over-generalization. For the optimal case, the number of misclassified points is 22.

Figure 3 presents the corresponding results for spread unary coding.

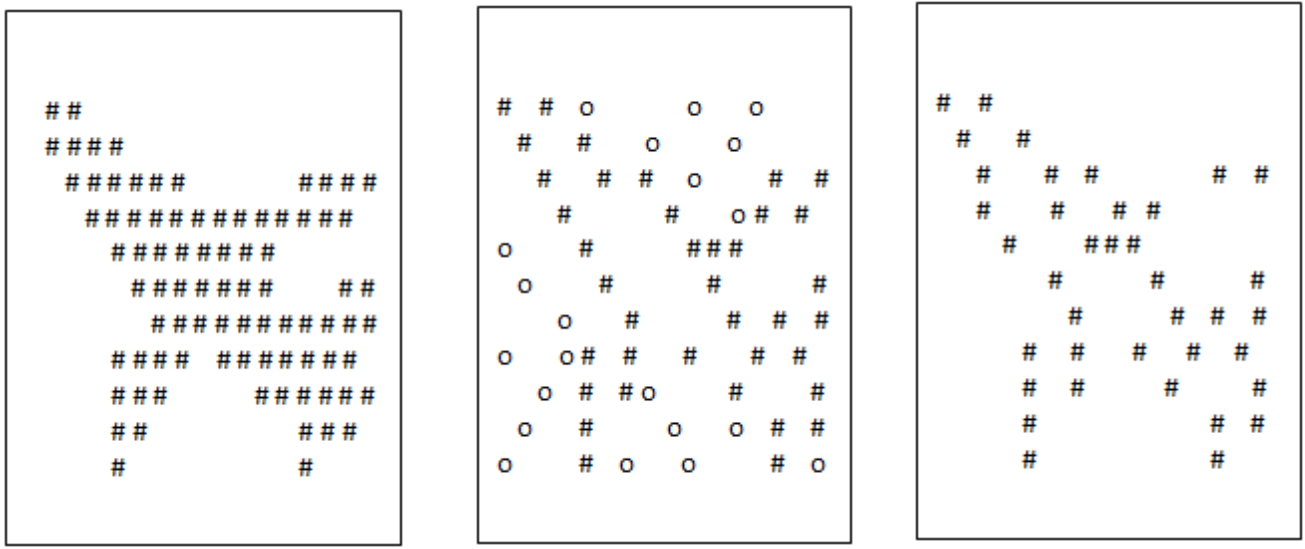

(a). Original Spiral Pattern  (b). Training Sample  (c). r = 1

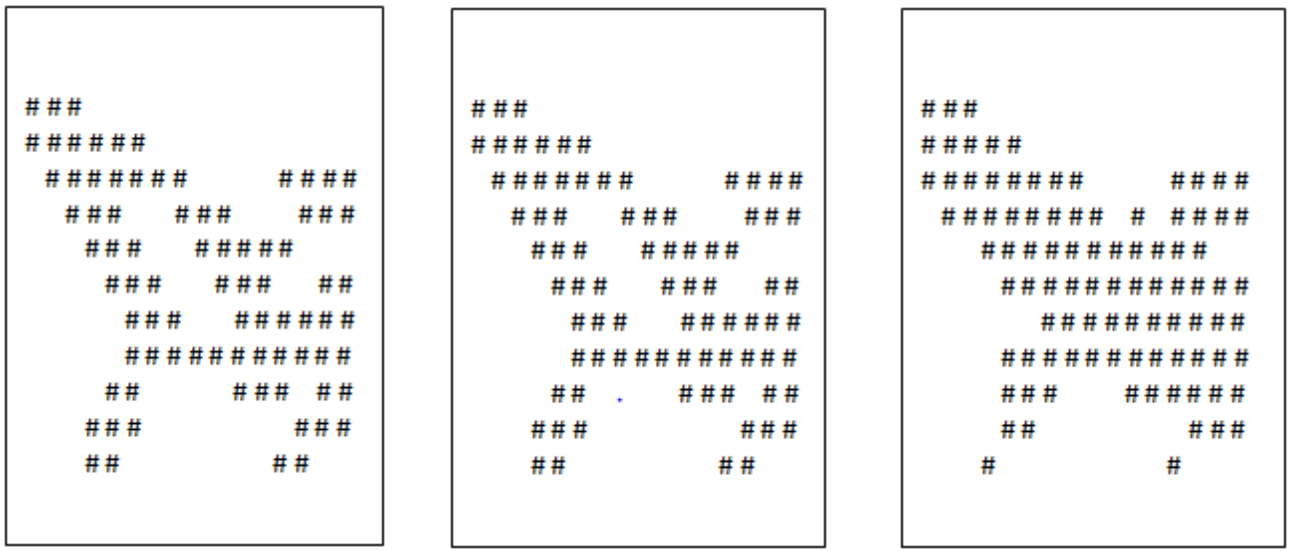

(d). r = 2  (e). r = 3  (f). r = 4

Figure 3: Results of Pattern classification using Spread Unary coding

Table 4: No. of points classified/misclassified in the pattern using unary coding

|  | r = 1 | r = 2 | r = 3 | r = 4 |
|---|---|---|---|---|
| Misclassified | 59 | 30 | 30 | 24 |
| Classified | 117 | 146 | 146 | 146 |

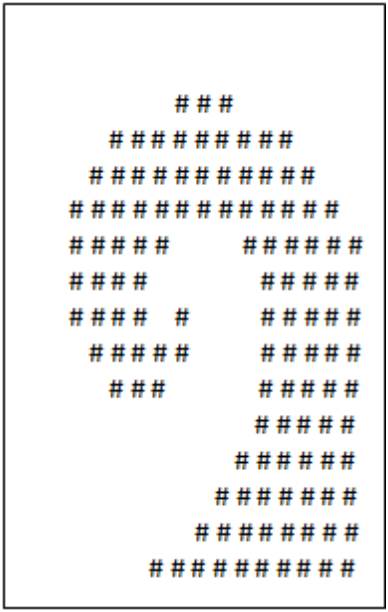
(a). Original Spiral pattern

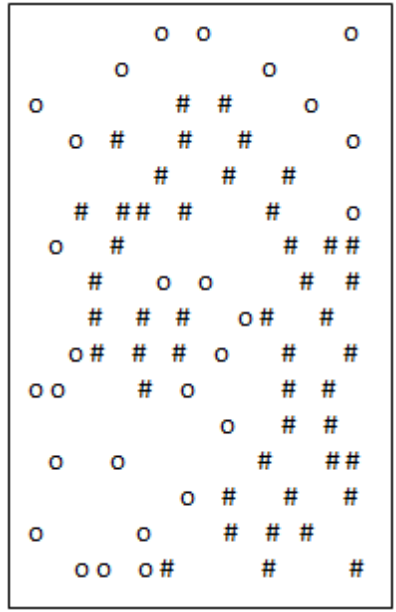
(b). Training sample

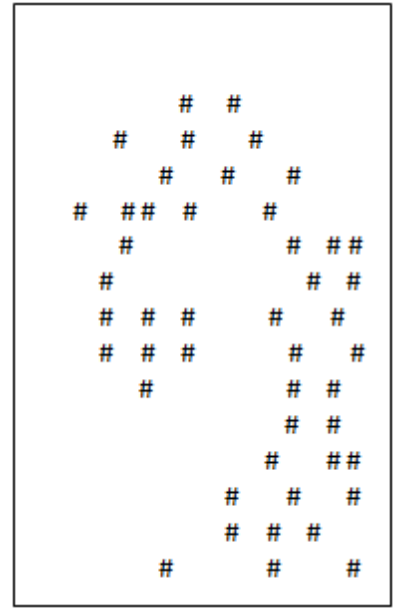
(c). r = 1

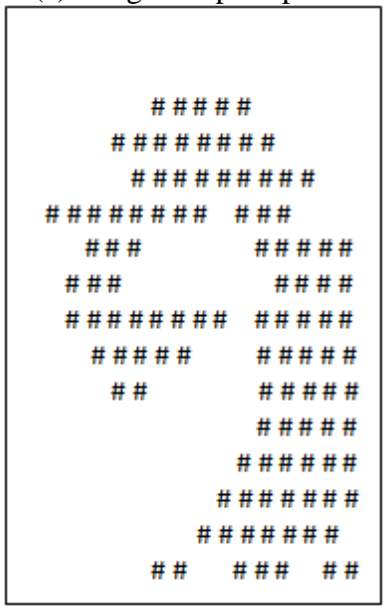
(d). r = 2

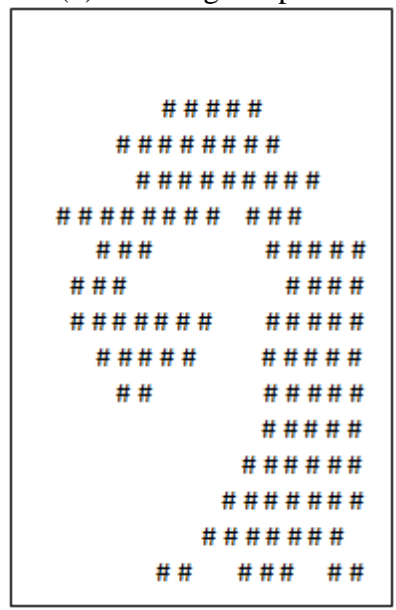
(e). r = 3

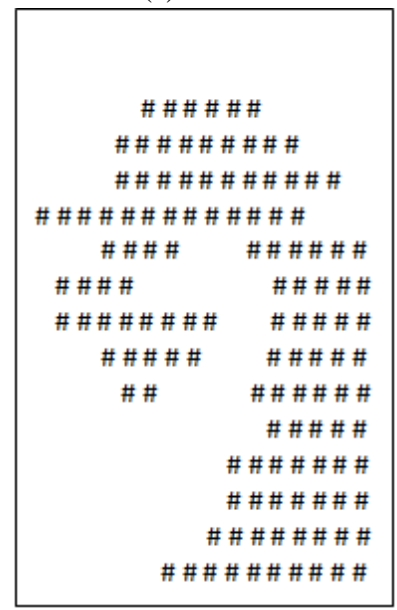
(f). r = 4

Figure 4: Results of Spiral Pattern classification using Spread Unary coding

The performance of spread unary coding is comparable in terms of the misclassified points for the best case when r=4. On the other hand, the case of classified points remains stable as we go from r=2 to r=4. Another example is provided in Figure 4 and Table 5

Table 5: No. of points classified/misclassified in the pattern using spread unary coding

|  | $r = 1$ | $r = 2$ | $r = 3$ | $r = 4$ |
| --- | --- | --- | --- | --- |
| Classified | 179 | 236 | 235 | 237 |
| Misclassified | 77 | 20 | 20 | 19 |

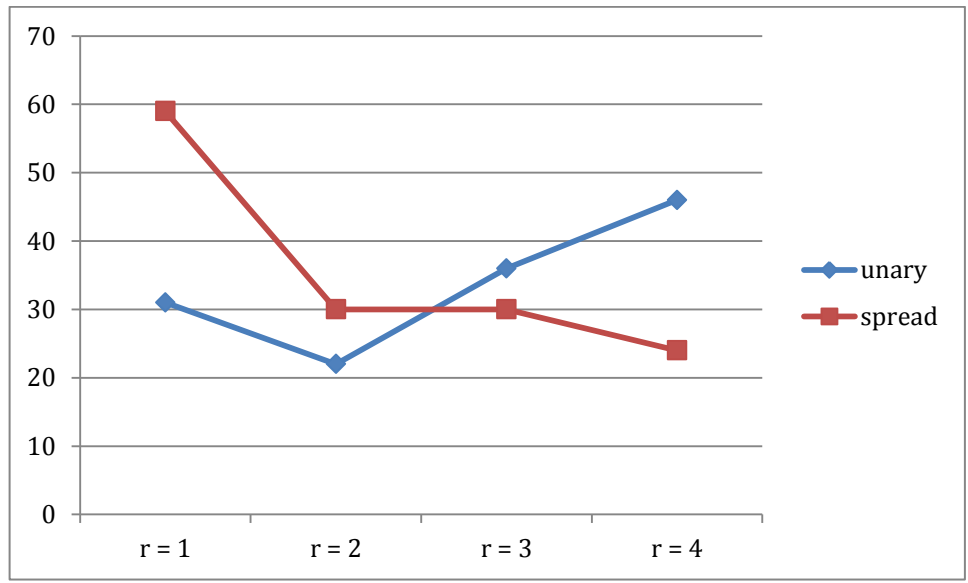

Figure 5: No. of misclassified points in unary vs spread unary

As shown in Figure 6, the number of misclassified points remains stable after the initial drop. This would make spread unary advantageous to use in many situations. Of course, we do not know if some version of CC4 is at the basis of biological learning. If it is, then this might be another reason why Nature has chosen it over other forms of neural network learning.

**Conclusions**
This paper adapted the corner classification algorithm (CC4) to train the neural networks using spread unary inputs. This is an important problem as spread unary appears to be at the basis of data representation in biological learning. The modified CC4 algorithm was tested using the pattern classification experiment and the results are found to be quite good. The improvement in performance on the use of spread unary coding rather than unary came in as a surprise.

Specifically, we show that the number of misclassified points is not particularly sensitive to the chosen radius of generalization. Since the choice of the radius depends on domain knowledge in the standard unary case and one can make errors in that choice, the relative stability of it in spread unary is a significant improvement. This also makes for an important new insight into the workings of the radius of generalization in CC4 networks.

**References**
1. A. Ponnath, Instantaneously Trained Neural Networks. Computer Research Repository, arXiv:abs/cs/0601129, 2006.
2. S. Kak,  A class of instantaneously trained neural networks. Information Sciences, vol. 148, 97-102, 2002.
3. S. Kak, New algorithms for training feedforward neural networks. Pattern Recognition Letters, vol. 15, pp. 295-298; 1994.
4. Z. Zhang et al., TextCC: New feed forward neural network for classifying documents instantly. Lecture Notes in Computer Science, Volume 3497, Jan 2005, Pages 232 – 237.
5. S. Kak, On generalization by neural networks. Information Sciences, vol. 111, pp. 293-302, 1998
6. S. Kak, On training feedforward neural networks. Pramana, vol. 40, pp. 35-42, 1993.
7. S. Kak, Y. Chen, L. Wang, Data mining using surface and deep agents based on neural networks. 16th Americas Conference on Information Systems. Lima, Peru, August 2010.
8. K.-W. Tang and S. Kak, Fast classification networks for signal processing. Circuits, Systems Signal Processing, vol. 21, pp. 207- 224, 2002.
9. K.-W. Tang and S. Kak, A new corner classification approach to neural network training. Circuits Systems Signal Processing, vol.17, pp. 459-469, 1998.
10. S. Kak, The three languages of the brain: quantum, reorganizational, and associative. In Learning as Self-Organization, K. Pribram and J. King (editors). Lawrence Erlbaum Associates, Mahwah, NJ, 1996, pp. 185-219.
11. I.R. Fiete, R.H. Hahnloser, M.S. Fee, and H.S. Seung. Temporal sparseness of the premotor drive is important for rapid learning in a neural network model of birdsong. J Neurophysiol., 92(4), pp. 2274–2282, 2004.
12. I.R. Fiete and H.S. Seung, Neural network models of birdsong production, learning, and coding. New Encyclopedia of Neuroscience. Eds. L. Squire, T. Albright, F. Bloom, F. Gage, and N. Spitzer. Elsevier, 2007.
13. S. Kak, Generalized unary coding. Circuits, Systems and Signal Processing, 2015.
14. S. Kak, The Architecture of Knowledge. New Delhi, Motilal Banarsidass, 2004.
15. J. Zhu and G. Milne, Implementing Kak neural networks on a reconfigurable computing platform. In FPL 2000, LNCS 1896, R.W. Hartenstein and H. Gruenbacher (eds.), Springer-Verlag, 2000, p. 260-269.
16. A. Shortt, J. G. Keating, L. Moulinier, C. N. Pannell , Optical implementation of the Kak neural network. Information Sciences vol. 171,  pp. 273-287, 2005.
17. Z. Jihan, P. Sutton, An FPGA implementation of Kak's instantaneously-trained, fast-classification neural networks. Proceedings of the 2003 IEEE International Conference on Field-Programmable Technology (FPT), December 2003.